\documentclass[11pt]{article}
\usepackage{mtsummit2019}
\usepackage{times}
\usepackage{url}
\usepackage{latexsym}
\usepackage[small,bf]{caption} 
\usepackage{arabtex}
\usepackage{utf8}
\usepackage[utf8]{inputenc}
\usepackage[T2A,LAE,T1]{fontenc}

\usepackage[text={16cm,24cm}]{geometry}

\newcommand{\hide}[1]{}

\newcommand{\smt} {SMT\tiny $_{tgt++}$}
\newcommand{\nmtST} {NMT\tiny $_{scr/tgt++}$}

\title{The Impact of Preprocessing on Arabic-English\\ Statistical and Neural Machine Translation}

\author{Mai Oudah, Amjad Almahairi$^\dagger$ and Nizar Habash\\
Computational Approaches to Modeling Language Lab \\
New York University Abu Dhabi, UAE\\
$^\dagger$Element AI, Canada\\
  {\tt \{mai.oudah,nizar.habash\}@nyu.edu}\\
  {\tt amjad.almahairi@elementai.com}}

\date{}

\begin{document}
\maketitle
\setarab
\novocalize

\setcode{utf8}
\begin{abstract}
Neural networks have become the state-of-the-art approach for machine translation (MT) in many languages. 
While linguistically-motivated tokenization techniques were shown to have significant effects on the performance of statistical MT, it remains unclear if those techniques are well suited for neural MT. 
In this paper, we systematically compare neural and statistical MT models for Arabic-English translation on data preprecossed by various prominent tokenization schemes. Furthermore, we consider a range of data and vocabulary sizes and compare their effect on both approaches.
Our empirical results show that the best choice of tokenization scheme is largely based on the type of model and the size of data. We also show that we can gain significant improvements using a system selection that combines the output from neural and statistical MT.
\end{list}
\end{abstract}

\section{Introduction}
Neural machine translation (NMT) has been rapidly attracting the attention of the research community for its promising results \cite{D141179,bahdanau-2014,wu16,vaswani17}. NMT is composed of two neural networks, an encoder and a decoder, where the encoder is fed a sentence from the source language and the decoder generates its translation, word by word, in the target language. Recently, NMT has been shown to outperform other MT systems in many language pairs, e.g. German-English, French-English and Basque-English \cite{W174725,hybrid17,lrec18}. While Arabic MT has been mostly developed under statistical MT (SMT), NMT has also been applied and studied recently \cite{HabashS06,almahairi2016first,DurraniDSV17}. 

Linguistically-motivated tokenization has shown to have a significant effect on SMT, particularly in the case of morphologically rich languages like Arabic \cite{HabashS06}. However, it remains unclear if such techniques are well suited for NMT, where language-agnostic tokenizations, e.g. byte-pair encoding (BPE) \cite{bpe}, are widely used.
\newcite{almahairi2016first} has looked into Arabic SMT and NMT, achieving the highest accuracy using the Penn Arabic Treebank (ATB) tokenization, with 51.2  and 49.7 BLEU points for SMT and NMT, respectively. 

In this paper, we study the impact of different preprocessing techniques in Arabic-English MT on both SMT and NMT, by examining various prominent tokenization schemes.
We conduct learning curve experiments to study the interaction between data size and the choice of tokenization scheme. We study the performance under morphology-based and frequency-based tokenization schemes, provided by MADAMIRA \cite{MADAMIRA2014} and BPE, respectively, on in-domain data. In addition, we evaluate the best performing models on out-of-domain data. Our results show that the utilization of BPE for SMT can be effective and allows achieving a good performance even with a small vocabulary size of 20K. Moreover, the results show that the performance of NMT is especially sensitive to the size of data. We notice that NMT suffers with long sentences, and thus, we utilize system selection, which yields significant improvements over both approaches. Our best system significantly outperforms previous results reported on the same in-domain test data by +4 BLEU points \cite{almahairi2016first}.

The rest of the paper is organized as follows. The related work is presented in Section~2. Section~3 describes our proposed approach. Section~4 illustrates the experimental settings. The results are reported in Section~5. In Section~6, we discuss our findings. Finally, we conclude the paper and mention the future work in Section 7.


\section{Related Work}

Many studies have compared the performance of different MT models on translation tasks \cite{lrec18,almahairi2016first,DurraniDSV17}. However, the data preprocessing was not unified across those models. For example, BPE is only applied to the training data utilized by the NMT system, but not SMT \cite{almahairi2016first,DurraniDSV17}.

\newcite{HabashS06} investigated and compared across some preprocessing schemes for Arabic, describing and evaluating different methods for combining them. The main preprocessing schemes were Simple Tokenization, Decliticization (degrees 1 to 3), and Arabic Treebank Tokenization. Decliticization of degree 2 outperformed the rest when applied individually. They reported improvement in MT performance when combining different schemes together.

\newcite{almahairi2016first} compared NMT and SMT on Arabic translation, and showed that NMT performs comparably to SMT. The best performance is achieved when Penn Arabic Treebank (ATB) tokenization is used with 51.19  and 49.70 BLEU points for SMT and NMT, respectively. 

The idea of system selection for MT exists in the literature, but mostly for model selection under the same approach (SMT or NMT) \cite{Devlin2012,SalloumEAHD14}. 


\section{Approach}
In our study, we systematically compare SMT and NMT on the following dimensions.
\subsection{Source Language Tokenization}
Much research has shown the importance of tokenization and orthographic normalization for SMT and NMT, as they deal with data sparsity \cite{el2012orthographic,HabashS06,zalmout-habash:2017}. Tokenization schemes can either be morphology-based or statistical/frequency-based \cite{MADAMIRA2014,bpe}. We investigate both in the context of Arabic MT, both separately and in combination, to observe their interaction. We normalize Alif `\RL{ا}' and Ya `\RL{ي}' in all schemes, where Hamza is removed from the variants of Hamzated Alif (e.g.`\RL{أ}',`\RL{إ}') to become `<a>', the Alif Maqsura `\RL{ى}' is replaced with Ya `\RL{ي}' and the diacritics are removed. 
 
 {\bf Morphology-based} This tokenization scheme relies on the linguistic rules of the source language. We explore three schemes under this category \cite{HabashS06,zalmout-habash:2017}:  1) Simple Tokenization (Raw) that splits off punctuation and numbers; 2) Penn Arabic Treebank (ATB) Tokenization, which splits all clitics except definite articles; 3) Decliticization (D3), which splits all clitics.
 
\begin{figure*}[h!tb]
	\hspace{-3pt}
	\centerline{\includegraphics[trim=0.0cm 22cm 0.0cm 2.5cm,clip,width=1.33
	\textwidth]{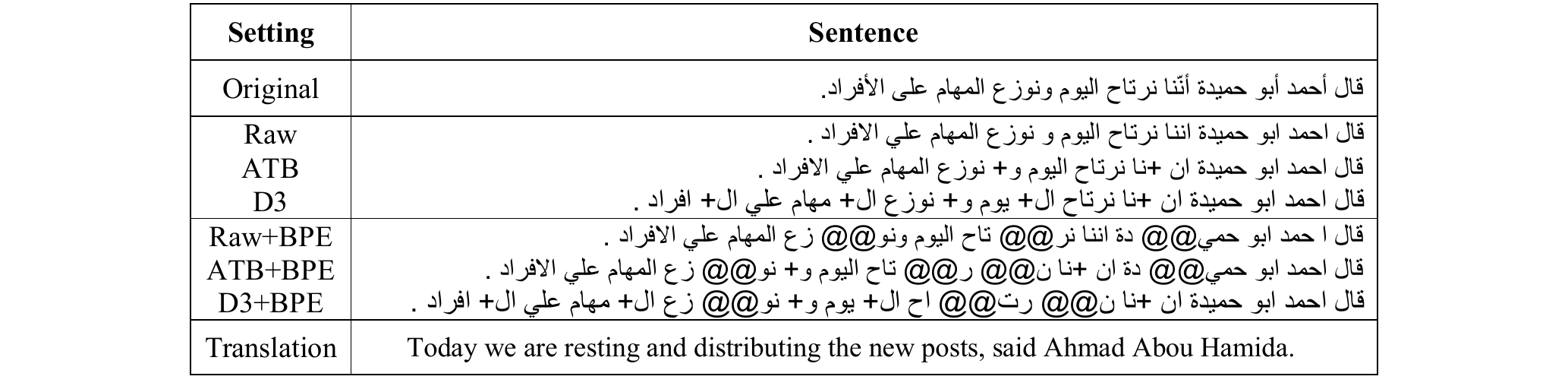}}
	\caption{Tokenization schemes applied to an example.}
	\label{tok-examples}
\end{figure*}

 {\bf Frequency-based} We use byte-pair encoding (BPE) \cite{bpe}, which is an iterative compression approach that replaces the most frequent pair of characters in a sentence with a unique sequence of characters. It allows for a fixed-size vocabulary representation. Figure \ref{tok-examples} shows an example across Raw and Tok schemes with/without BPE on top.

\subsection{Training Data Size}
We conduct a learning curve experiment to explore how much both Arabic-English SMT and NMT can benefit from adding more training data with each tokenization scheme. \newcite{HabashS06} have conducted a similar learning curve study for SMT.
Each tokenization scheme may result in a different number of tokens per sentence; hence, a sentence-length filter will discard more sentences from more verbose schemes. This would lead to some schemes having access to more words than others. 
Therefore, we adopt \newcite{el2012orthographic}'s approach of filtering training parallel data based on the D3 scheme as a reference scheme for selecting sentences of length up to 100 tokens. Thus, the same sentences will be selected across different tokenization schemes. 

\subsection{Target Language Resources}
 We design the training so that both systems will have access to the same additional target language resources besides the target side of the training parallel corpus. In SMT, target language resources are used to build language models for fluency improvement. Whereas, many works have proven pretrained word embeddings to be useful in neural network models \cite{Qi:2018}, and therefor, the same additional TLR are used to learn pretrained word embeddings that support the decoder in NMT. 
Here, the $_{tgt++}$ designation next to the system name indicates the use of additional TLR. 

\subsection{Input Length and System Selection}
Many have reported NMT performing worse with long sentences \cite{W14-4012,Koehn:2017}, which was caught in our error analysis and thus we explored combining the two MT systems via a system selection approach, where the selection of either translation is based on which is closer to the input length as a criterion. Whereas the sentence BLEU score is the criterion in the Oracle system selection.

\section{Experimental Settings}
\begin{figure*}[h!tb]
	\centerline{\includegraphics[trim=0.0cm 21.80cm 0.0cm 2.7cm,clip,width=1.15\textwidth]{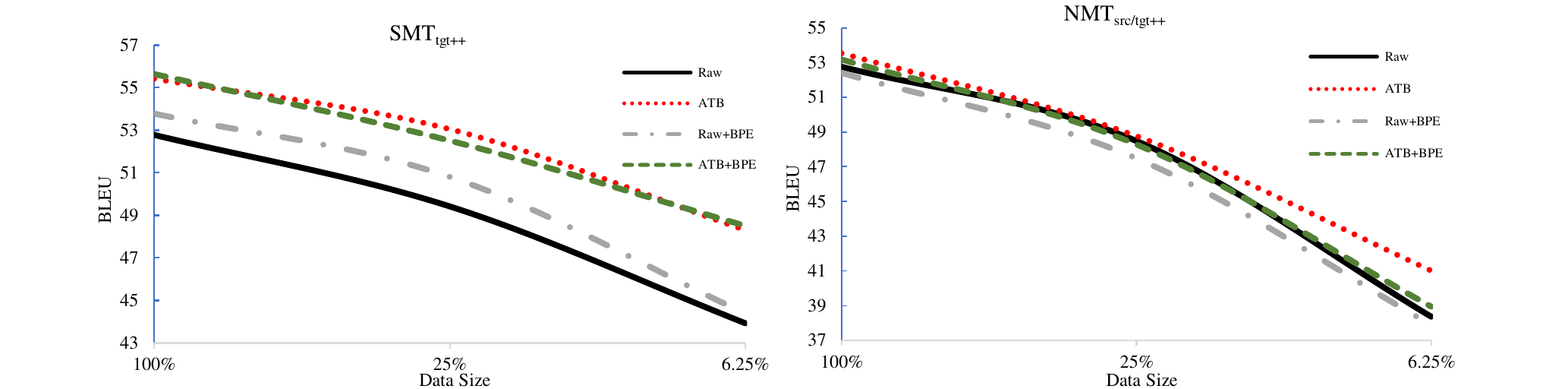}}
	\caption{The performance on in-domain test (MT05) under different settings with different training data sizes.}
	\label{Learning_curve_plots}
\end{figure*} 
\begin{table*}[h!]
\setlength\tabcolsep{3pt}
\begin{center}
\scalebox{1}{
\begin{tabular}{|c||c||c||c||c||c||c|}
\hline {} & \bf  \#Vocab & \bf {\smt} & \bf CI & \bf {\nmtST} & \bf CI & \bf $P$-value
 \\ \hline
Raw & 331K & 52.78 & $\pm$ 0.98 & 52.76 & $\pm$ 1.24 & 0.412 \\
ATB & 208K & 55.42 & $\pm$ 1.07 & \bf 53.54 & $\pm$ 1.20 & 0.002 \\
D3 & 190K & 54.66 & $\pm$ 1.02 & 53.51 & $\pm$ 1.20 & 0.027 \\ \hline
Raw+BPE & 20K & 53.78 & $\pm$ 1.10 & 52.41 & $\pm$ 1.17 & 0.003 \\
ATB+BPE & 20K & \bf 55.64 & $\pm$ 1.11 & 53.18 & $\pm$ 1.15 & 0.001 \\
D3+BPE & 20K & 54.59 & $\pm$ 1.07 & 53.38 & $\pm$ 1.16 & 0.018 \\ \hline
\end{tabular}
}
\end{center}
\caption{\label{mt-results} Comparing Raw, ATB and D3 Tokenized cases without/with BPE on in-domain test (MT05), in terms of BLEU scores, where the Confidence Interval (CI) and $P$-value are reported. Bold font highlights best results by SMT and NMT. }
\end{table*}

\subsection{Datasets}
The training dataset contains 1.2M sentence pairs in newswire (NW) domain from three Linguistic Data Consortium (LDC) resources: LDC2004T18, LDC2004T14, and LDC2007T08. For tuning, we use LDC2010T12 (MT04), which consists of 1,075 sentence pairs in NW and government documents. As for the in-domain testing, we use LDC2010T14 (MT05), which consists of 1,056 sentence pairs in NW, and has four English reference translations. We look into the performance of the systems in out-of-domain data using LDC2014T02 (MT12), which consists of 1,535 sentence pairs mostly web collection, and has four English reference translations.

\subsection{Preprocessing} 
MADAMIRA \cite{MADAMIRA2014} is utilized for morphology-based tokenization of the source side. \newcite{bpe}'s BPE implementation is used for learning and applying BPE models. We set vocabulary size to 20K in BPE learning after exploring multiple vocabulary sizes, including 10K, 20K and 30K, where the 20K setting achieved comparable results to the 30K and outperformed the 10K. Each BPE model is trained on source side of training data of the respective experiment. While Moses' \cite{Koehn07b} tokenizer and lowercaser are used for preparing the target side. 

\subsection{SMT settings}
We use Moses 3.0 \cite{Koehn07b} to train SMT models with maximum phrase length of 8 tokens. Two versions of the language model are examined: 1) trained solely on the target side of the training dataset, and 2) trained on the target side and the English Gigaword 5\textsuperscript{th} edition. 

\subsection{NMT settings}
We use the encoder-decoder with the \textit{general} global attention architecture as introduced by \newcite{Luong-2015}.
All the NMT models have been trained using OpenNMT toolkit \cite{opennmt} with no restriction on input's vocabulary.
We use long short-term memory units (LSTM) \cite{Hochreiter:1997:LSM}, with hidden units of size 500 and two layers in both the encoder and decoder. The word embedding vector size for source/target is 300. 

English pretrained word embeddings were trained as skip-gram model \cite{DBLP:journals/corr/abs-1301-3781} via gensim tool \cite{rehurek_lrec_gensim} with settings: (size=300, window=8, min count=5) on English Gigaword 5\textsuperscript{th} edition \cite{Graff2003eng} dataset. Arabic embeddings were trained on the Arabic Gigaword 5\textsuperscript{th} edition \cite{LDC:Gigaword-5} via FastText \cite{fasttext}, which showed better performance with morphologically rich languages \cite{Erdmann-etal-2018}. We give the designation of $_{src/tgt++}$ to the system that uses both embeddings.


\subsection{Evaluation Metrics}
The evaluation results are reported in case insensitive BLEU scores \cite{BLEU} with their confidence intervals (CI) and $p$-values. Bootstrap resampling is used to compute statistical significance intervals \cite{Koehn_04_EMNLP}.

\begin{figure*}[h!tb]
	\hspace{-3pt}
	\centerline{\includegraphics[trim=0.2cm 11.5cm 0.0cm 2.5cm,clip,width=1.05\textwidth]{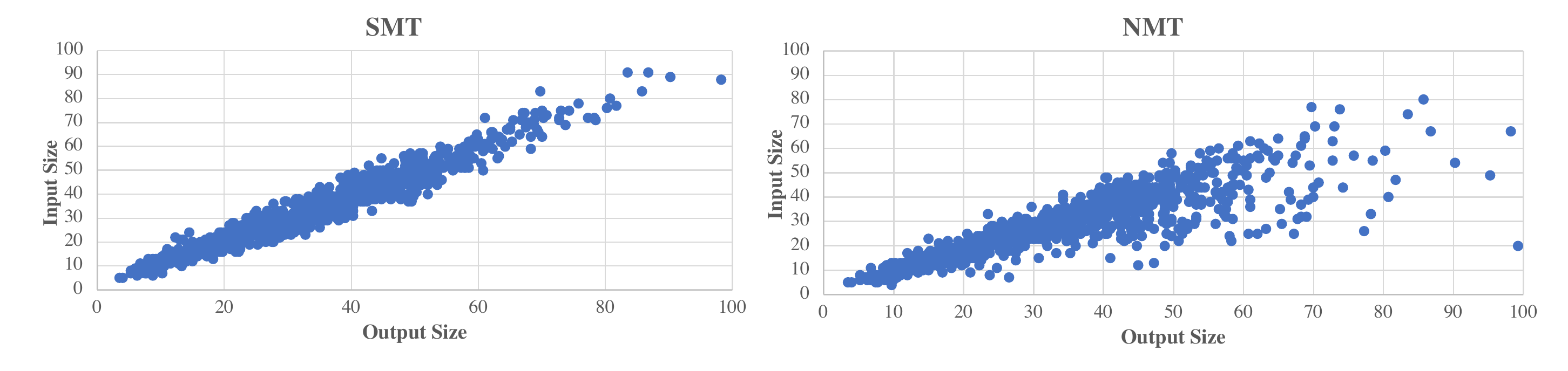}}
	\caption{The input size vs. output size in SMT and NMT, respectively, on MT05 with ATB tokenization. We notice that in NMT parts of the input sentences are dropped and not translated at all, which motivates the length-based selection.}
	\label{smt-nmt-chart}
	\vspace{5pt}
\end{figure*}

\begin{table*}[t!]
\setlength\tabcolsep{3pt}
\begin{center}
\begin{tabular}{|c|c|c|c||c|c|}
\hline  \multicolumn{2}{|c|}{\bf \smt }  & \multicolumn{2}{c||}{\bf \nmtST} & \bf System Selection & \bf Oracle \\ \hline
Setting &  BLEU &  Scheme &  BLEU &  BLEU &  BLEU \\ \hline
ATB+BPE & 55.64 & ATB & 53.54 &  \bf  56.18 & 61.26  \\ \hline
\end{tabular}
\end{center}
\caption{\label{sys-sel-results} BLEU score of the length-based system selection (using best models of SMT and NMT) when applied on in-domain test (MT05). }
\end{table*}

\begin{table*}[t!]
\setlength\tabcolsep{3pt}
\begin{center}
\begin{tabular}{|c|c|c|c||c|c|}
\hline \multicolumn{2}{|c|}{\bf \smt}  & \multicolumn{2}{c||}{\bf \nmtST} &    \bf System Selection & \bf Oracle \\ \hline
Scheme &  BLEU & Scheme & BLEU & BLEU & BLEU \\ \hline
ATB+BPE & 35.11 & ATB & 36.56 & \bf 37.96 & 39.11 \\ \hline
\end{tabular}
\end{center}
\caption{\label{blindtest-results} 
BLEU score of the length-based system selection (using best models of SMT and NMT) when applied on out-of-domain test (MT12).
}
\end{table*}

\begin{figure*}[h!tb]
	\hspace{-3pt}
	\centerline{\includegraphics[trim=0.0cm 2.5cm 0.0cm 2.0cm,clip,width=1.15\textwidth]{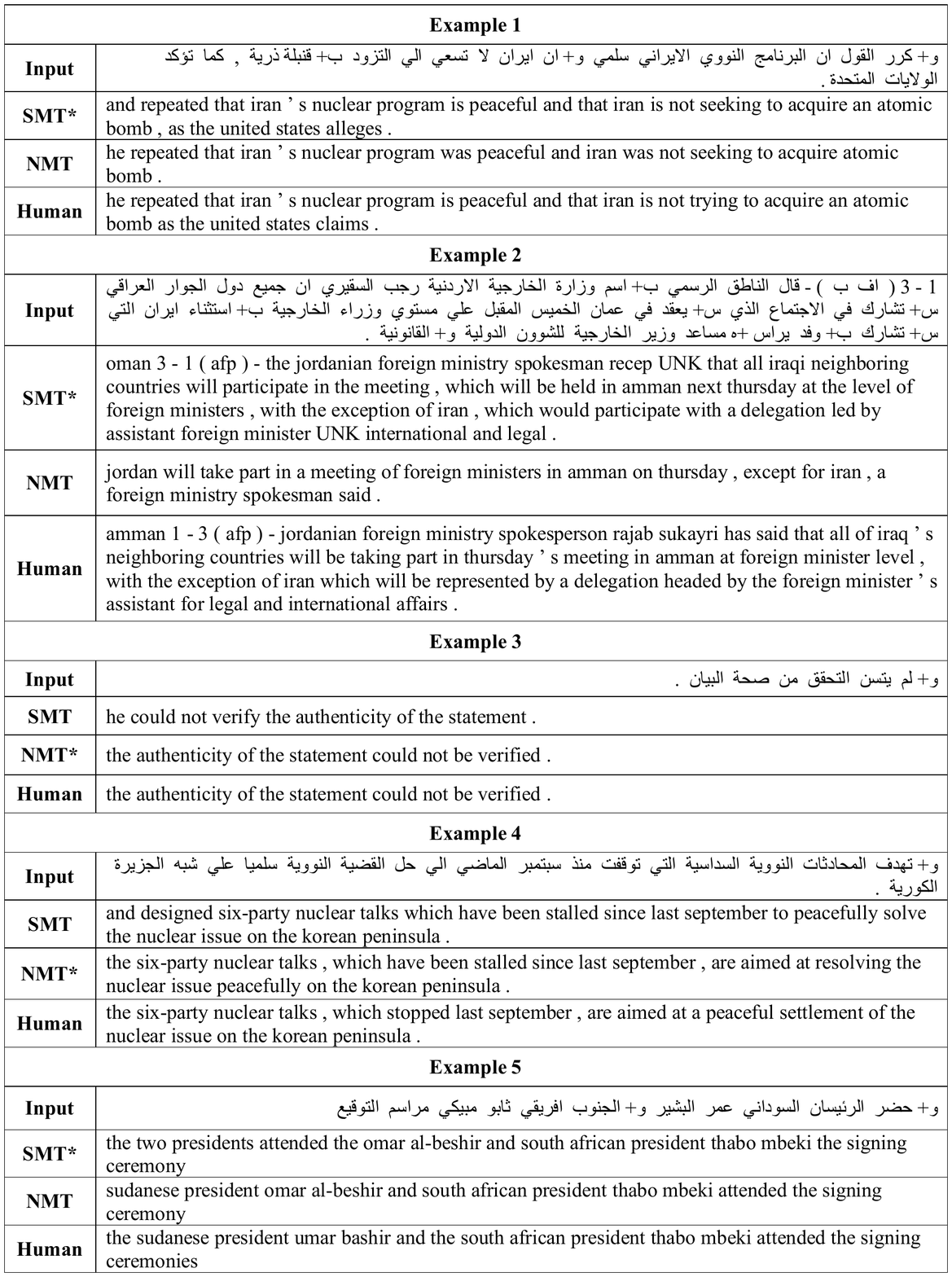}}
	\caption{Examples from MT05, with SMT and NMT outputs when ATB is used as a scheme. The *  designation next to the system name indicates the decision of the system selection. }
	\label{sys-selection-examples}
	\vspace{5pt}
\end{figure*}

\section{Results}
\subsection{Preprocessing and Learning Curve}
We examine Raw, ATB and D3 with and without BPE applied on top, across a learning curve where smaller sets of our training data (1.2M) are considered at $25$\% (300K) and $6.25$\% (75K) tokens. Figure \ref{Learning_curve_plots} illustrates the learning curve results for Raw (baseline) and ATB (overall best), with and without BPE. Figure \ref{Learning_curve_plots} shows the importance of training data availability, especially for NMT, and also that BPE impact can be seen in both systems, which we find interesting. Moreover, SMT is shown to be far more sensitive to preprocessing than NMT. 

Table \ref{mt-results} shows the best systems' results when $100$\% of the training data is tokenized by Raw, ATB and D3, with and without BPE on top of it, for SMT and NMT. It shows ATB+BPE and ATB to achieve the best results for SMT and NMT, respectively, which we find interesting as BPE is usually associated with NMT. The $p$-value indicates whether the difference between SMT and NMT results under the same tokenization scheme is statistically significant or not. The statistical significance is illustrated with $p$-value $<0.05$. 
{\smt} and {\nmtST} have comparable results at the baseline.
As expected, using more data for LM produces better results as well as the increase in training data size. 
Using pretrained word embeddings for both languages improve the NMT results significantly compared to only target ones by two BLEU points. As shown in Table \ref{blindtest-results}, the best {\nmtST} model (using ATB) outperforms {\smt} model (using ATB+BPE) by 1.5 BLEU against MT12 in the out-of-domain testing. 

\subsection{Error analysis}
Error analysis has shown that NMT output is more fluent than SMT's, especially with short sentences ($<50$ tokens), in contrast to long sentences where coverage and accuracy drop, which support related work. Figure \ref {smt-nmt-chart} shows dropping in NMT output size as the input size increases, especially $<40$ tokens, while SMT keeps more consistent output to the input size. 
So we explore system selection based on the closeness to the input length as well as Oracle results, where the selection is based on the highest output BLEU score. 
Tables \ref{sys-sel-results} and \ref{blindtest-results} show the results of length-based system selection on best models in SMT and NMT when applied to in-domain test (MT05) and out-of-domain testing (MT12), respectively, which illustrate improvements over the original BLEU scores.

Figure \ref{sys-selection-examples} illustrates five examples, where either SMT or NMT output is selected based on the output length compared to the source input size. 
In Example 1, the SMT output is selected over the NMT one as the NMT system drops the phrase after the comma and only translates the part before, however, fluently. Example 2, which represents much longer sentence, SMT output is selected over NMT, which translates the saying and drops the rest of the sentence. On the other hand, in Examples 3 and 4, NMT output is selected over SMT's for being the closest to the source input in terms of length. Furthermore, Example 5 represents a case where the system selection approach fails to select the better prediction (in terms of BLEU score) for the final output based on the source-output length comparison.

\section{Discussion}
We notice that morphology-based tokenization schemes improve the performance of MT systems regardless of the MT approach, but in different levels. The difference in scheme choice is less impactful on NMT; compared with SMT. The improvement range for NMT is 1.13 BLEU, while for SMT the range is 2.86 BLEU. While Raw results are almost the same for SMT and NMT; ATB improves both NMT and SMT; but the improvement is higher for SMT. Adding BPE helps SMT, while lowering vocabulary size. The effect of BPE on NMT is insignificant, which is a surprising result since BPE is often associated with NMT. 
Also, we significantly improve on \newcite{almahairi2016first}'s results by more than three BLEU points. 

Length-based system selection improves over both NMT and SMT results in in-domain and out-of-domain cases, significantly in the later, which indicates a hybrid MT system may be promising. Moreover, the huge jump in performance with Oracle selection shows that there is still room for potential improvement in system designs, for better accuracy and fluency. More TLR allow for better results in MT systems. When both Arabic and English pretrained word embeddings are used, the performance improves by more than two BLEU points compared to English only. 

\section{Conclusion and Future Work}
In this paper, we study the impact of various preprocessing techniques to Arabic-English MT under SMT and NMT, where  various  prominent tokenization schemes are examined. 
We conduct a learning curve analysis of the different preprocessing settings with incremental training data size, where ATB scheme performs consistently well along the learning curve. Moreover, we implemented a length-based system selection to deal with NMT's struggle with short sentences, and significant improvements. 
The empirical results show that the choice of tokenization scheme can be optimized based on the type of model to train and the data available. We also gain significant improvements using length-based system selection that combines the output from neural and statistical MT.
Our results significantly outperform the ones reported in the prior work when applied to in-domain test (MT05). As future work, we plan to examine training data of general domain with linguistically-motivated tokenization schemes to study further their impact on NMT under different neural models. Also, exploring sophisticated system selection schemes for potential improvement.

\section*{Acknowledgments}
The support and resources from the High Performance Computing Center at New York University Abu Dhabi are gratefully acknowledged.

\bibliographystyle{mtsummit2019}
\bibliography{ALLBIB-2.5,extra}

\end{document}